\documentclass[final]{article}

\usepackage[final]{nips_2017}

\usepackage[utf8]{inputenc} 
\usepackage[T1]{fontenc}    
\usepackage{hyperref}       
\usepackage{url}            
\usepackage{booktabs}       
\usepackage{amsfonts}       
\usepackage{nicefrac}       
\usepackage{microtype}      
\usepackage{amsmath}
\usepackage{code}

\DeclareMathOperator{\Update}{update}

\title{Provenance and Pseudo-Provenance for Seeded Learning-Based Automated Test Generation}

\author{
Alex Groce\\
School of Informatics, Computing, and Cyber Systems\\
Northern Arizona University\\
Flagstaff, AZ 86011
\And
Josie Holmes\\
SICCS\\
Northern Arizona University\\
Flagstaff, AZ 86011
}

\begin{document}

\maketitle

\begin{abstract}
  Many methods for automated software test generation, including some that
  explicitly use machine learning (and some that use ML more broadly
  conceived) derive new tests from existing tests (often referred to
  as seeds).  Often, the seed tests
  from which new tests are derived are manually constructed, or at least
  simpler than the tests that are produced as the final outputs of
  such test generators.  We
  propose annotation of generated tests with a \emph{provenance} (trail) showing
  how individual generated tests of interest (especially failing tests) derive
  from seed tests, and how the population of generated tests relates
  to the original seed tests.  In some cases, post-processing of
  generated tests can invalidate provenance information, in which case
  we also propose a method for attempting to construct
  ``pseudo-provenance'' describing how the tests \emph{could} have
  been (partly) generated from seeds.
\end{abstract}

\section{Seeded Automated Test Generation}

Automatic generation of software tests, including (security) fuzzing
\cite{aflfuzz,TrailBitsSeeded}, random testing \cite{csmith,ICSEDiff,Pacheco},
search-based/evolutionary testing \cite{FA11},
and symbolic or concolic execution
\cite{Whitebox,GodefroidKS05,KLEE,Person:2011:DIS:1993498.1993558,Marinescu:2012:MTS:2337223.2337308,issta14}
is essential for improving software security and reliability.  Many of
these techniques rely on some form of learning, sometimes directly
using standard algorithms
\cite{last2004artificial,Raffelt:2007:DTV:1787497.1787516,ARTChen,Groce:2002:AMC:646486.694482}
such as reinforcement learning \cite{ISSRE,ISOLA12,ReinforceBook}, and sometimes in a more broadly conceived
way.  In fact, using Mitchell's classic definition of machine learning
as concerning any computer program that improves its performance at some
task through experience \cite{Mitchell}, almost all non-trivial automated test generation
algorithms are machine-learning systems, with the following
approximate description:

\begin{enumerate}
\item Based on results of running all past tests ($T$), produce a new test
  $t' = f(T)$ to execute.
\item Execute $t'$ and collect data $d'$ on code coverage, fault detection and other
  information of interest for the execution of $t'$.
\item $T = \Update(T,t',d')$
\item Go to step 1.
\end{enumerate}

Performance here (in Mitchell's sense) is usually measured by
collective code coverage or fault detection of tests in $T$, or may be
defined over only a subset of the tests (those deemed most useful,
output as a test suite).  The function $f$ varies widely: $f$ may
represent random testing with probabilities of actions determined by
past executions \cite{AndrewsL07}, a genetic-algorithms approach where
tests in $T$ are mutated and/or combined with each other, based on
their individual performances
\cite{McMinn04search-basedsoftware,FA11,aflfuzz}, or an approach using
symbolic execution to discover new tests satisfying certain
constraints on the execution path \cite{Whitebox,GodefroidKS05,KLEE}.
A similar framework uses reinforcement learning, but constructs each
test on-the-fly and performs $\Update$ calls after every step of
testing \cite{ISSRE}.
A common feature however, is that many methods do not begin with the
``blank slate'' of an empty $T$.  Instead, they take as an initial
input a population of tests that are thought to be high-quality (and,
most importantly, to provide some guidance as to the structure of
valid tests), and proceed to generate new tests from these \emph{seed}
tests
\cite{aflfuzz,Person:2011:DIS:1993498.1993558,Marinescu:2012:MTS:2337223.2337308,issta14,STVR_seeding}.
Seed tests are usually manually generated, or tests selected from a
previously generated suite for their high coverage or fault detection
\cite{YooHarman,stvrcausereduce}.  It is generally the case that seed
tests are more easily understood by users than newly generated tests.
For example, seed tests often include only valid inputs and
``reasonable'' sequences of test actions, while generated tests, to
improve coverage and fault detection, often include invalid inputs or
bizarre method call sequences.

For example, consider the extremely popular and successful American
Fuzzy Lop (AFL) tool
for security fuzzing \cite{aflfuzz}.  It usually begins fuzzing
(trying to generate inputs that cause crashes indicating potential
security vulnerabilities) from a corpus of ``good'' inputs to a
program, e.g., actual audio or graphics files.  When a corpus input is mutated and the result is
``interesting,'' by a code-coverage based heuristic, the new input is
added to the corpus of tests to use in creating future tests.  Many
tools, considered at a high level, operate in the same fashion, with the
critical differences arising from engineering aspects (how tests are executed
and instrumented), varied heuristics for selecting tests to mutate,
and choice of mutation methods.  AFL records the origin of each test
in its queue in test filenames, which suffices in AFL's case because
each test produced is the result of a change to a single, pre-existing
test, in most cases, or the merger of two tests, in rarer cases.

This kind of trace back to the source of a generated test in some seed
test (possibly through a long trail of also-generated tests) is
essentially a \emph{provenance}, which we argue is the most easily
understood explanation of a learning result for humans, in those cases
(such as testing) where the algorithm's purpose is to produce
novel, interesting objects from existing objects.

This simple approach used in AFL works for cases where the provenance of a test is
always mediated by mutation, making for a clear, simple ``audit
trail.''  However, a more complex or fine-grained approach is
required when the influence of seeds is probabilistic, or a test is
composed of (partial) fragments of many tests.  Moreover, AFL provides
no tools to guide users in making use of what amounts to an internal
book-keeping mechanism, and does not produce provenance output designed for human
examination.
Finally, tests, once generated, are frequently manipulated in ways
that may make provenance information no longer valid:  a test produced
from two seed tests (or seed test-derived tests) may be reduced
\cite{DD} so that one of the tests no longer is present at all, for example.

In this paper, we propose to go beyond the kind of simple mechanisms
found in AFL, and
offer the following contributions:

\begin{itemize}
\item We present an implementation of provenance for an algorithm that involves
  generating new tests from partial sequences from many seed tests.
\item We discuss ways to present information about not just the
  provenance of a single test, but the impact on future tests of
  initial seed tests.  While single-test provenance is useful for
  developers debugging, information on general impact of seeds is more
  important for design and analysis of test generation configuration
  and algorithms.
\item We identify test manipulations that partially or completely
  destroy/invalidate provenance information, and propose an algorithm
  for producing a \emph{pseudo-provenance}, showing how the tests generated
  \emph{could} have been generated from seeds, even if they were not actually
  thus generated, and discuss abstractions that enable pseudo-provenances.
\end{itemize}

\section{A Simple Seeded Generation Algorithm with Provenance}

We implemented a novel test generation technique for the TSTL
\cite{tstlsttt,nfm15,issta15,tstl} test generation language and tool for
Python.  In this approach, the seed tests are split into (usually short)
sub-sequences of length $k$.  In place of the usual algorithm for
random testing, where a new test action is randomly chosen at each
step during testing, our approach always attempts to follow some
sub-sequence, in a best-effort fashion (if the next step in the
current sub-sequence is not enabled, it is skipped).  When a test
generated in this fashion covers new code (the usual metric for
deciding when to learn from a test, in such methods), it too is broken into sub-sequences
and the sequences are added to the sub-sequence pool and used in generation of
future tests.

In TSTL, a test is a sequence of components (test actions), and the
provenance of a test generated using this algorithm involves numerous
tests, and varying parts of those tests.  We extended TSTL to allow
every component of a test to be annotated with a string.  Whenever a
component from a sequence in the sequence pool is added to a test, it
is labeled with the file name of the source test and the exact
location in that test of the component.  Figure \ref{fig:example}
shows part of a high-quality test for an AVL data structure library.  In the
example, we first generated a set of ``quick tests''
\cite{icst2014,stvrcausereduce}, small tests that together obtain
maximum coverage.  We then used sequences from these tests to guide
testing, with the goal of producing a single test with maximal code
coverage (the highest coverage from any one quick test is 173 branches
and 131 statements).  The complete generated test of which the first fragment
is shown in Figure \ref{fig:example}
covers 204 branches and 152 statements.  Each step (component) in the
test is labeled with its exact source in one of the 6 seed tests.
Because all tests generated (including new quick tests enhancing
coverage) are thus annotated, the source of a test component can always be
traced back to an initial seed test.  Here, the file names of the
tests are not highly informative; however, using a recently-proposed
technique for automatically giving generated tests high-quality names
\cite{Daka:2017:GUT:3092703.3092727}, the information could be even
more useful.  In practice, many seed tests would also be named for
previously detected faults they are associated with, thus providing
considerable information as to the context of test components.

\begin{figure}
{\scriptsize
\begin{code}
int1 = 13                   \# STEP 0   ;;; quick1.test:11
int0 = 7                    \# STEP 1   ;;; quick1.test:14
int2 = 16                   \# STEP 2   ;;; quick2.test:4
avl1 = avl.AVLTree()        \# STEP 3   ;;; quick5.test:3
avl1.insert(int2)           \# STEP 4   ;;; quick0.test:15
avl1.insert(int1)           \# STEP 5   ;;; quick0.test:16
avl0 = avl.AVLTree()        \# STEP 6   ;;; quick3.test:1
int1 = 10                   \# STEP 7   ;;; quick3.test:2
avl0.insert(int0)           \# STEP 8   ;;; quick3.test:3
avl0.insert(int1)           \# STEP 9   ;;; quick3.test:4
avl0.delete(int0)           \# STEP 10  ;;; quick3.test:5
avl1.insert(int2)           \# STEP 11  ;;; quick5.test:10
int2 = 14                   \# STEP 12  ;;; quick5.test:11
\end{code}
}
\caption{Example test generated with fine-grained provenance information}
\label{fig:example}
\end{figure}

In this example algorithm, provenance is certain and direct:  each component of a
test arises from one previously-generated or seed test.  However, in
some cases (e.g., learned models) multiple tests may influence the
\emph{probability} of a component appearing.  For example, the
probability of each possible component could be non-linearly proportional to how
many seed (and learned) tests that component appears in.  In such cases,
however, the same approach applies, except that instead of a single
source, each component is labeled with a set of contributing tests,
along with their degree of contribution (e.g., if a component appears
5 times in some seed tests, it will increase probability of generating
that component more than a seed test in which it only appears once).

\section{Presenting Collective Provenance Information}

While provenance of components of a single test is the most
interesting information for debugging and understanding a newly
generated test, the most difficult and frequently performed part of an
automated testing effort is the \emph{effort to understand the overall
behavior of the test generation}.  For that task, information beyond
the provenance of single tests is essential.

Fortunately, the critical information is easily summarized.  Usually
simply tabulating frequency of a seed test contributing to a generated
test is sufficient, either at the level of the entire test or, for an
algorithm like the one presented above, at the level of individual
components of a seed test.  For example, if we apply our new algorithm
to test {\tt pyfakefs} \cite{pyfakefs}, a popular Python file system
simulator for testing, we discover that a test that first creates,
then removes, a sequence of nested directories is the single most
useful test to mine for new coverage, out of a set of 50 seed tests
covering basic file system functionalities.  However, more useful
perhaps is an analysis of actual components used, abstracted to their
general type of operation (what file system call is involved).  This
shows that, in order, the most important operations for exposing new
coverage are:  {\tt symlink}, {\tt makedirs}, {\tt rename}, and {\tt
  close}, followed by variations of {\tt open}.  The file-level data
combined with this data shows that our seeds provide very few
variations on the creation and destruction of multiple directories at
once, making the one test with such a sequence extremely important,
despite these not being the most contributory components for adding
test coverage.  At this point, adding more seeds with the
``over-used'' components is a plausible next step for improving test
generation.

\section{Test Manipulations and Pseudo-Provenance}

The approaches presented above are applicable to a large set of test
generation methods.  However, tests generated are often manipulated
after generation.  In particular, they are very often \emph{reduced}
in size \cite{DD}, producing a test with fewer components that
preserves either fault detection or code coverage \cite{icst2014}
properties of the original test.  Reduction usually preserves
provenance in a sense (components not removed maintain their
provenance annotations), but a long sequence is likely to be broken up
and have its context lost.  Understanding the provenance after
reduction is therefore likely harder, because there are fewer long sequences
from existing tests.  Moreover, test \emph{normalization} \cite{Groce:2017:OTR:3092703.3092704}, which uses
term rewriting to change (vs. simply remove) test components tends to
destroy provenance information completely (any components rewritten
during normalization
lose their provenance, since the new component is produced by a brute
force search, not from seed tests).

However, we can still provide some of the information that provenance
would have provided, by showing how an algorithm based on replaying
sub-sequences of seed tests \emph{could} have produced the new tests.
This applies to both a simple sub-sequence-replay approach such as
described above, or to more complex approaches, such as a Markov model
of the seed tests.

\subsection{A Greedy Pseudo-Provenance Algorithm}

The algorithm for constructing a pseudo-provenance is simple.  We take
the set of seed tests, and a test for which no provenance exists (or
for which provenance has been partially destroyed).  First, for each
component of the test missing provenance, we compute all possible
positions in all seed tests compatible with that component.  Normally,
this will be matching components only, but we also allow optional abstraction
(as with provenance summaries) to the \emph{type} of test action,
rather than the specific action (since normalization and alpha
conversion tend to modify exact variable names from source tests).
For components that already have provenance, the set of compatible
seed components is just the actual provenance source.   After this, we
iterate through the test, at each step removing any compatible
provenances that do not extend a previous sub-sequence.  Whenever the
set of compatible components from seeds becomes empty, the current
sub-sequence cannot be extended, so we return to the previous step,
and construct a pseudo-provenance by iterating backwards from that
position until we encounter test components already labeled with a
provenance, annotating each step with the same arbitrarily selected
provenance source from possible sequences in seeds (since all are guaranteed to produce an equally sized
sub-sequence).

The problem is similar to computing a string alignment (e.g., as in
computing Levenshtein distances \cite{lev}), except that there are
multiple strings to potentially align with.  While in principle a more
expensive algorithm than our greedy approach could produce better
pseudo-provenances, we believe that making initial sub-sequences as
long as possible, in order to ``orientate'' readers is more
important in debugging, and the exact quality is not critical for summarization
purposes.

Using our greedy approach, we can construct the provenance of the test
in Figure \ref{fig:example} exactly, with the exception that in
several cases we find a better matching sequence than was really
responsible for the generated test.  For example, {\tt STEP 3} now is
associated with {\tt quick0.test:14} rather than {\tt quick5.test:3}.
This is also an actual way the test could have been produced by our
sequence-based generation
algorithm, and in many cases would make understanding the relationship
to a seed test easier.  Pseudo-provenance may be useful even in cases
where real provenance exists, since for our generation algorithm, at least (or a
Markov-based approach), a pseudo-provenance is equally valid in the
sense that it provides a causal explanation of the generated test.
The pseudo-provenance may relate to a less likely sequence of events,
but it may also relate to a more likely sequence of events.  In many
cases, the ability to understand generated tests by having extended
sub-sequences of seed tests is more important than such questions of
generation-method probability.

\section{Conclusions}

In this paper, we present the problem of explaining the origins of
generated tests derived from seed tests, and propose and implement (in
TSTL \cite{tstlsttt}, with code on GitHub \cite{tstl}) approaches for
handling this problem in practical testing.  One unusual feature of automated test
generation is that even when provenance is lost in either generation
or post-processing of tests, constructing a pseudo-provenance is still
useful in understanding the context of generated tests, and how they
relate to seed tests.



\bibliographystyle{plain}

\end{document}